\pdfoutput=1

\documentclass[11pt]{article}

\usepackage{acl}

\usepackage{times}
\usepackage{latexsym}

\usepackage[T1]{fontenc}

\usepackage[utf8]{inputenc}

\usepackage{microtype}

\usepackage{hyperref}       
\usepackage{url}            
\usepackage{booktabs}       
\usepackage{amsfonts}       
\usepackage{nicefrac}       
\usepackage{xcolor}         
\usepackage{graphicx}
\usepackage{algorithm}
\usepackage{algorithmic}
\usepackage{soul}
\usepackage{amsmath}
\usepackage{bbm}
\usepackage{amsthm}
\usepackage{subfigure}
\usepackage{multirow}
\usepackage{verbatim}
\usepackage{comment}
\usepackage{booktabs}
\usepackage{tablefootnote}
\usepackage{hyphenat}

\title{Interpretable Research Replication Prediction \\ via Variational Contextual Consistency Sentence Masking}

\author{Tianyi Luo$^1$, Rui Meng$^{2}$, Xin Eric Wang$^{1}$, Yang Liu$^{1}$ \\
  $^1$Computer Science and Engineering, University of California, Santa Cruz \\
  $^2$Lawrence Berkeley National Laboratory, University of California, Berkeley\\
  {\tt \{tluo6, xwang366, yangliu\}@ucsc.edu} \\
  {\tt rmeng@lbl.gov}
  \\}

\begin{document}
\maketitle
\begin{abstract}
Research Replication Prediction (RRP) is the task of predicting whether a published research result can be replicated or not. Building an interpretable neural text classifier for RRP promotes the understanding of why a research paper is predicted as replicable or non-replicable and therefore makes its real-world application more reliable and trustworthy. However, the prior works on model interpretation mainly focused on improving the model interpretability at the word/phrase level, which are insufficient especially for long research papers in RRP. Furthermore, the existing methods cannot utilize a large size of unlabeled dataset to further improve the model interpretability. To address these limitations, we aim to build an interpretable neural model which can provide sentence-level explanations and apply weakly supervised approach to further leverage the large corpus of unlabeled datasets to boost the interpretability in addition to improving prediction performance as existing works have done. In this work, we propose the \underline{V}ariational \underline{C}ontextual \underline{C}onsistency \underline{S}entence \underline{M}asking (\textbf{VCCSM}) method to automatically extract key sentences based on the context in the classifier, using both labeled and unlabeled datasets. Results of our experiments on RRP along with European Convention of Human Rights (ECHR) datasets demonstrate that VCCSM is able to improve the model interpretability for the long document classification tasks using the area over the perturbation curve and post-hoc accuracy as evaluation metrics. 
\end{abstract}

\section{Introduction}

\begin{figure}[!h]
\centering
\includegraphics[width=7.8cm,height=6.22cm]{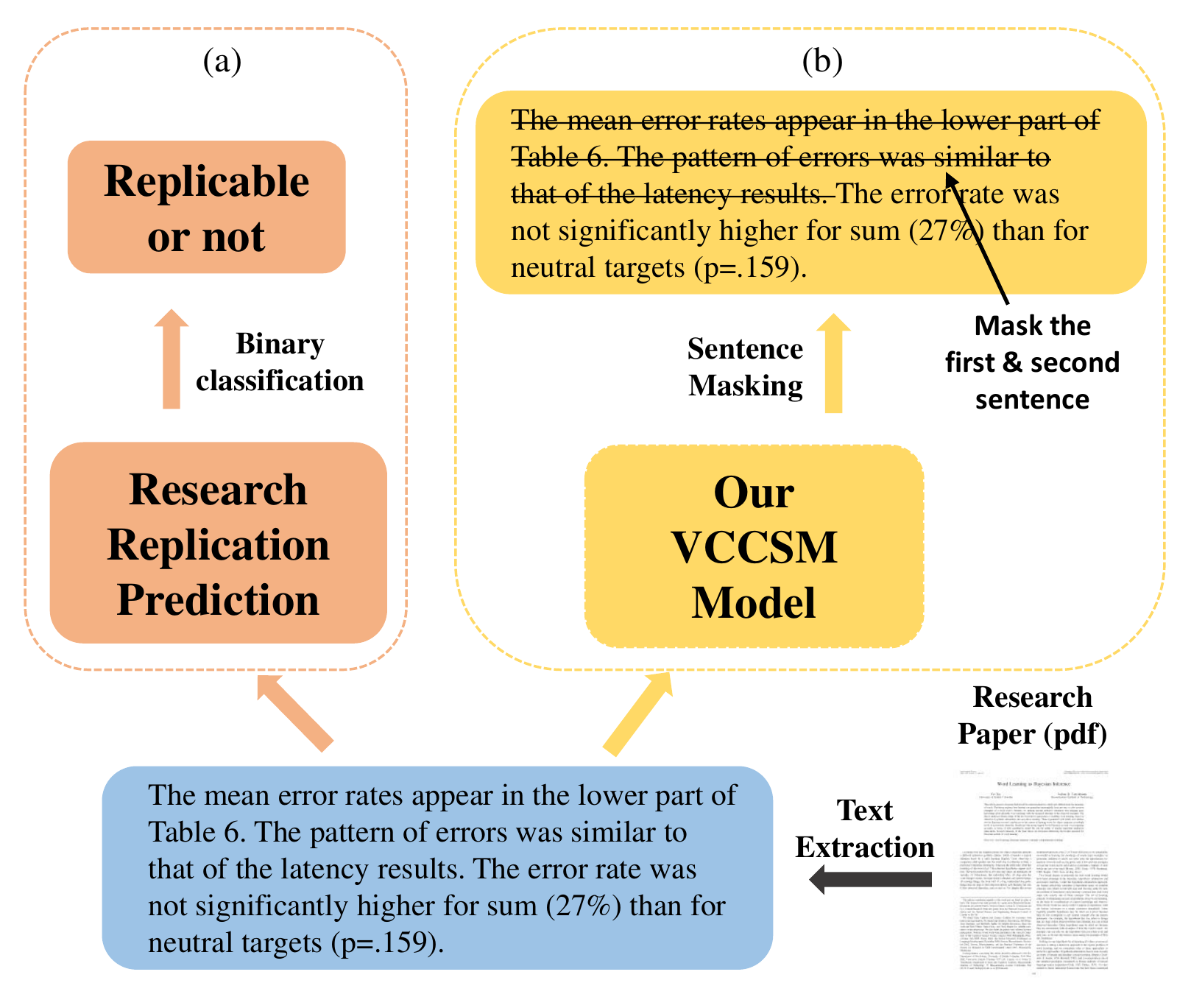}
\caption{(a) Given the text information of a research paper, Research Replication Prediction (RRP) task predicts whether the paper can be reproduced or not. (b) Having the same input as (a), our VCCSM model can keep the important sentences (through masking unimportant ones) which are related to reproducibility. 
}
\label{fig:illustration_mask}
\vspace{-0.4cm}
\end{figure}

Scientific research results that cannot be reproduced are unreliable and negatively impact the development of science. Therefore, it is important to know whether a published research result can be replicated or not. To this end, domain researchers have conducted several direct replication projects in contemporary published social science studies \cite{camerer2016evaluating, camerer2018evaluating, ebersole2016many, klein2018many, open2015estimating}. Such direct replication, however, is very time-consuming and expensive. A much more efficient and cheaper alternative, Machine Learning (ML), is utilized for predicting research replication \cite{dreber2018predicting,Yang2019,altmejd2019predicting,luo2020research}. 
In this paper, we model the task of predicting research replication as a binary classification problem and name it Research Replication Prediction (RRP) task which is shown in Figure \ref{fig:illustration_mask}(a).
Nonetheless, applying the neural network models in the context of RRP faces two challenges. The first challenge is that the existing neural network models used in RRP are characterized as a black box because their predictions are hardly understandable. Without intelligible explanations for the predictions, results of RRP may not be widely accepted as reliable and trustworthy. Despite the progress in interpretable machine learning 
\cite{hechtlinger2016interpretation, smilkov2017smoothgrad,singh2018hierarchical,serrano2019attention, han2020explaining,chen2020learning,chen2021explaining}
, the existing works mostly focus on improving the interpretability only at the word/phrase level which might work well for short documents. However, research papers in our RRP problem are usually lengthy (the average length of words is about 10,000). Building interpretable models for long documents is a challenging task due to the massive amount of textual information.

The second salient challenge is the small size of labeled dataset in RRP due to the high cost (e.g., funding requirement, human labor, etc.) of direct replications. Training an interpretable neural network also requires a large size of labeled dataset and 
weakly supervised learning can help utilize the unlabeled dataset. 
Although weakly supervised approaches have been utilized to make use of a large size of unlabeled dataset \cite{berthelot2019mixmatch, xie2019unsupervised, chen2020mixtext}, they have mainly focused on improving the prediction performance but not the interpretability. We therefore aspire to build a weakly supervised interpretable neural text classifier for predicting research replication that can leverage the existence of the large corpus of unlabeled articles to boost up both the prediction performance and the interpretability.

To tackle the first challenge mentioned above, we built an interpretable neural network model which can automatically select key sentences instead of words/phrases by adding a \emph{variational sentence masking} layer on the input layer which is a simple modification of network architecture but can effectively improve the model interpretability.  

By adding a \emph{variational sentence masking}, we can adopt information bottleneck framework \cite{tishby2000information,alemi2016deep} to train the model and improve both the prediction performance and interpretability by identifying important sentences.  
In addition, we hypothesize that whether to mask a sentence or not should also depend on its context (whether other sentences in the same paper are masked) in the case for long research papers because the information provided by extracted key sentences should 
not be redundant. Therefore, we invoke a contextual sentence masking approach using the LSTM model \cite{hochreiter1997long}. The extracted key sentences after masking are considered as our interpratable outcomes for each research paper. 

To resolve the second challenge, we developed a new weakly supervised method which makes use of unlabeled dataset to improve both the prediction performance and interpretability. Specifically, we adopted the consistency training methods \cite{laine2016temporal,tarvainen2017mean,xie2019unsupervised}
which regularize model predictions to be invariant to the small noises added to the input.
Consistency training were used to improve the prediction performance with the help of unlabeld dataset \cite{xie2019unsupervised}. In this paper, to improve the interpretability along with the prediction performance, we propose a consistency training method with sentence masking through replacing the noises-added input of unlabeled dataset by masked sentences. Specifically, for each unlabeled research paper, we generate the first prediction by using only the extracted key sentences 
after the sentence masking. Then, we generate the second prediction using all the sentences in the research paper without masking. The consistency check is then imposed upon the two predictions by minimizing the difference between them. Through the consistency training, an extra large size of unlabeled dataset can be utilized to make model continually learn how to extract the key sentences of a research paper so that the model interpretability is further improved.

In sum, our main contribution is the proposal of a variational contextual consistency sentence masking (VCCSM) method as shown in Figure \ref{fig:illustration_mask}(b) that is able to (1) extract the key sentences based on the context of a research paper and (2) leverage the large number of unlabeled sets of papers using a consistency checking mechanism. We present experimental results to validate the usefulness of our proposed methods on two neural network models, LSTM \cite{hochreiter1997long} and BERT \cite{devlin2018bert} on the RRP along with ECHR datasets. In particular, we find VCCSM is able to improve both the replication prediction accuracy and the interpretability for long research papers and general long documents.

\section{Related Work}

\paragraph{Blackbox Research Replication Prediction} Research Replication Prediction, knowing whether a published research result is replicable or not, is important. Recently, several large scale of direct replication projects have been conducted in social science studies to alleviate the replication crisis. But the cost of direct replication is too high to have a large size of annotated dataset. Therefore, an alternative ML method that is much cheaper and more efficient than direction replication is utilized in RRP. \citet{luo2020research} proposed a neural text classifier to achieve the best performance on RRP. But their model is a blackbox and cannot provide faithful explanations about why a research paper is predicted as replicable or non-replicable.

\paragraph{Interpreting Neural Networks} Various approaches have been proposed to interpret neural network models from the post-hoc manner, such as gradient-based \cite{simonyan2014deep,hechtlinger2016interpretation,sundararajan2017axiomatic}, attention-based \cite{serrano2019attention}, decomposition-based \cite{murdoch2018beyond,singh2018hierarchical}, 
example-based methods \cite{koh2017understanding,han2020explaining}, and word masking \cite{chen2020learning}. However, these interpretation methods have their own limitations, including only work with specific neural network model, render doubts on faithfulness, and need additional work to provide the explanations based on trained models. In this paper, we focus on model-agnostic explanation methods. More specifically, we follow the research of masking methods which can improve both the prediction performance and interpretability by adopting information bottleneck framework \cite{tishby2000information,alemi2016deep} to identify important sentences.

\paragraph{Improving interpretability via word masking}
\citet{chen2020learning} proposed a word masking method which can automatically select important words in the training process and build interpretable neural text classifiers by formulating their problem in the framework of information bottleneck. The proposed solution mainly deals with the short text and the average length (words) in all the seven datasets they used are less than 300. Four of them are less than 25. In constrast, the average length (words) of research papers in our RRP task is about 10,000 which is much longer than the ones used in \cite{chen2020learning}. Therefore, we view word masking as insufficient for our task. On the other hand, \citet{chen2020learning} learn independently on whether each word is masked or not. But context matters, especially for long documents. Different from prior work, we utilized the context information (whether other sentences in the same paper are masked or not) of each sentence by applying LSTM models to decide whether to mask this sentence or not. We hypothesize that context masking is better than independent masking, especially for long documents such as the research papers in RRP.

\paragraph{Consistency Training on Unlabeled Dataset}

The annotated data in RRP is collected using direction replication and its size is small. Therefore, weakly supervised learning methods need to be used to improve the model performance in RRP with the help of the unlabeled dataset. The existing weakly supervised methods applied in RRP focus mainly on improving the prediction performance, but less so about the model interpretability. 

Consistency training can improve the robustness of models by regularizing model predictions to be invariant to small noise applied to input examples \cite{sajjadi2016regularization,clark2018semi}. 
\citet{xie2019unsupervised} proposed to substitute the traditional noise injection methods in the consistency training with high quality data augmentations so that a new consistency training based weakly supervised method is proposed and the performance is improved with the help of unlabeled dataset. But they focused only on improving the prediction performance. 

In this paper, we conduct the consistency training on the unlabeled dataset to improve both prediction performance and interpretability by substituting the traditional noise injection methods with sentence masking methods, which is the major contribution of our paper. More specifically, we first mask the unimportant sentences and keep the critical sentences. Then we make the predictions on the kept key sentences the same as the ones based on all the sentences in the research paper without masking. Finally, we conducted the consistency check by minimizing the difference between them.

\section{Problem Statement}

In this paper, our main goal is to improve the intepretibility of neural textual classifier for Research Replication Prediction (RRP). First we introduce the RRP task.

\paragraph{Research Replication Prediction (RRP) task} In RRP, we hope to build a model $f$ that takes each research \texttt{article} as input and predicts whether the made research claim is replicable or not $f(\texttt{article}) \in \{0~ \text{(non-replicable), 1~\text{(replicable)}}\}$. There are different definitions and criteria for claiming a research paper to be replicable. In this work, a research paper is replicable means that an independent replication can provide evidence of a statistically significant effect in the same direction as the original paper.

\paragraph{Interpretable Research Replication Prediction} In this paper, we aim to build an interpretable neural textual classifier for RRP. Improving the model interpretability can help us understand why a research paper is predicted as replicable or non-replicable and make its application in the real world achieve more reliability and trustworthiness. Different from generating post-hoc explanations based on well-trained models, we adopt the information bottleneck framework \cite{tishby2000information,alemi2016deep} to train our model and build a more interpretable neural textual classifier for RRP.

\paragraph{Preliminaries and notations} To perform the above task, we have an labeled training dataset $\mathcal L:=\{(x_i, y_i)\}_{i=1}^{L} $, an unlabeled dataset $\mathcal U:=\{x_i\}_{i=1}^{U} $, and a test dataset $\mathcal T:=\{(x_i, y_i)\}_{i=1}^{T} $, where $L, U,$ and $T$ are the number of labeled training, unlabeled training, and testing datasets respectively. 
$x_i$ contains a sequence of sentences $x_i = [x_{i1}, x_{i2},...x_{ij}...,x_{iS}]$ in the $i_{\text{th}}$ research paper and $S$ is the maximum number of sentences in a research paper in RRP task. For the $j_{\text{th}}$ sentence in $x_i$, $x_{ij}=[x_{ij1}, x_{ij2},...x_{ijk}...,x_{ijK}]$, where $n$ is the maximum number of words in a sentence and $x_{ijk} \in \mathbb{R}^{d}$ which indicates the word embeddings as the model input. All the sentences have the same length $K$ by truncating. And $y_i$ is $x_i$'s binary classification label which is either `1' (replicable) or `0' (non-replicable). A neural textual classifier can be trained to output the replication labels given any new research paper $x_i$.

\section{Method}
The details of our proposed variational contextual consistency sentence masking (VCCSM) method are described in this section.

\subsection{Model Overview}

\begin{figure*}[!ht]
\centering
\includegraphics[width=12cm,height=6.2cm]{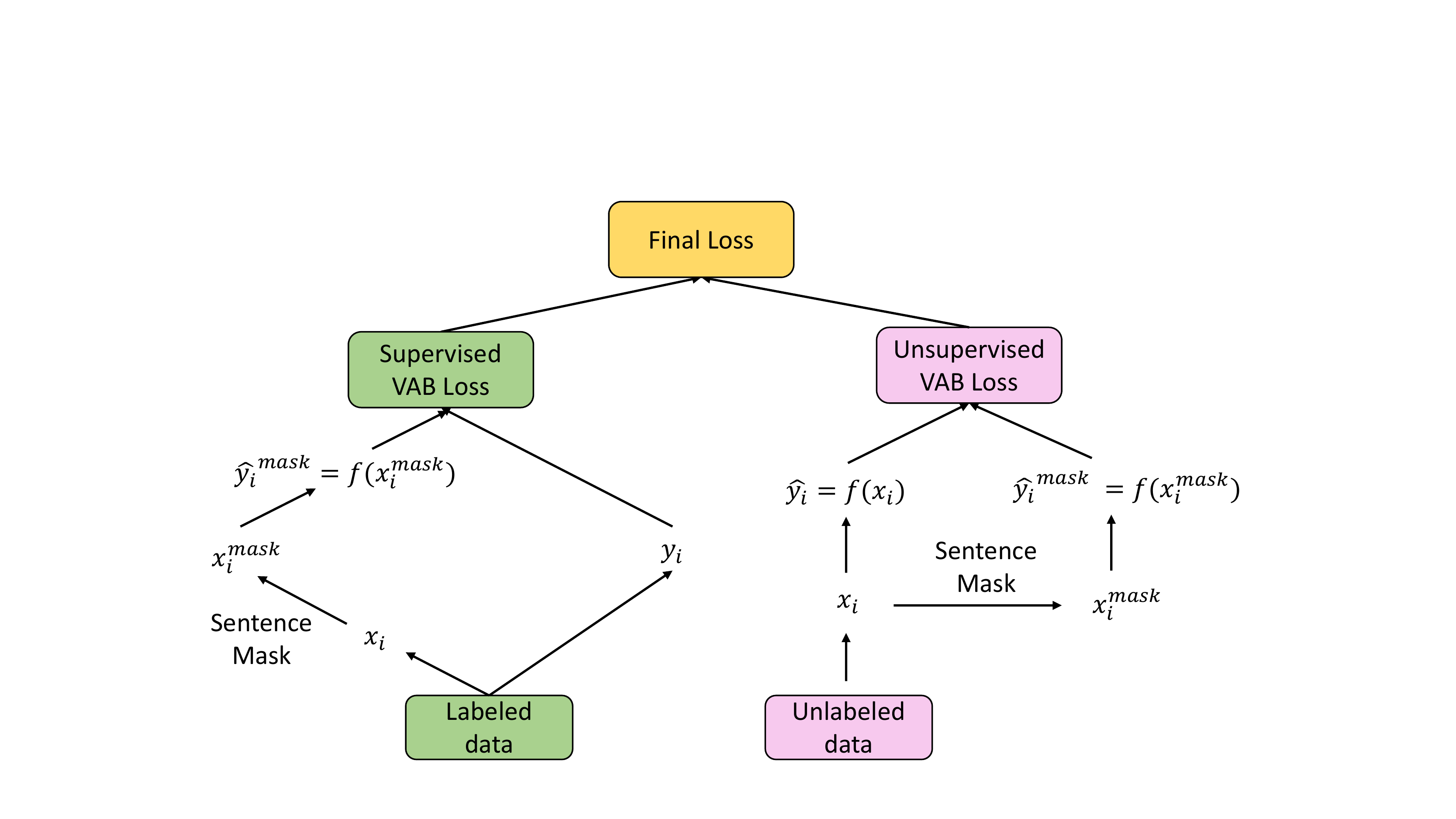}
\vspace{-0.1cm}
\caption{The architecture of variational contextual consistency sentence masking (VCCSM).
}
\label{fig:cssm_architecture}
\vspace{-0.2cm}
\end{figure*}

Our model contains two key modules: variational contextual sentence masking and consistency training. Variational contextual sentence masking module 
is applied in the training on both labeled and unlabeled datasets. Consistency training is only used in the training on the unlabeled dataset.

In the training on labeled dataset, variational contextual sentence masking module extracts the key sentences via contextual masking (LSTM model). Then the supervised loss is calculated and optimized to minimize the difference between prediction using the extracted sentences as the input and the ground truth label in the information bottleneck framework. The formula of supervised loss will be described later in this section and the architecture of the model on how to train the labeled dataset is shown in the left part of Figure \ref{fig:cssm_architecture}.

In the training on unlabeled dataset, different from the prior works, we conduct the consistency training on the unlabeled dataset to improve both prediction performance and interpretability by substituting the traditional noise injection methods with sentence masking methods. 
Consistency training can improve the model robustness by regularizing model predictions to be invariant to small noise applied to input examples \cite{sajjadi2016regularization,miyato2018virtual,clark2018semi}. Typical noise injection methods included additive Gaussian noise, dropout noise or adversarial noise. 
The existing consistency training based work e.g., \cite{xie2019unsupervised} focuses only on improving the prediction performance instead of interpretability. The consistency training methods utilized in this paper are based on variational contextual sentence masking and can also improve the model interpretability. Our optimization goal is to minimize the difference between prediction using the extracted vital sentences
and prediction made on all the sentences without masking in the information bottleneck framework. The formula of unsupervised loss will be described later in this section and the architecture on how to train the unlabeled dataset is shown in the right part of Figure \ref{fig:cssm_architecture}.

\subsection{Variational Contextual Sentence Masking}

Inspired by \citet{chen2020learning},
we want to add a mask layer $M$ after the sentence embeddings layer to help the model select the key sentences, where $M=[M_{1}, M_{2},...M_{j}..., M_{S}]$ and $S$ is the maximum number of sentences in a research paper. 
The embedding of each sentence is concatenated by word embeddings included in this sentence. 

Each $M_{j} \in \{0, 1\}$ is a binary random variable to decide whether we mask this sentence or not. For each sentence in one research paper, $M_{j}$ should be related to both the current sentence and the sentences around it (context). Therefore, we use LSTM model to generate the contextual sentence mask $M_{j}$ for the $j_{\text{th}}$ sentence in one research paper, where 
. $x$ can be any given research paper. This contextual sentence mask layer $M$ together with the sentence embeddings are considered as the input of neural network text classifiers in RRP, which is denoted as follows,
\begin{flalign}
\label{eq:x_mask}
Z=X^{\text{mask}}=M\bigodot X,
\end{flalign}
where $\bigodot$ is an element-wise multiplication, $X$ are all the examples in any given dataset, and $X^{\text{mask}}$ denotes the internal representations of all the examples. Our goal is to optimize $M$ so that the model can extract the key sentences for each research paper.

The information bottleneck theory aims to learn an encoding $Z$ of the input $X$ with maximal information on predicting the target $Y$ while keeps $X$'s the least redundant information \cite{tishby2000information,alemi2016deep}. As proven effective and flexible in identifying important features \cite{chen2020learning}, the information bottleneck framework is employed in our model. 
we want to make $Z=X^{\text{mask}}$ maximally expressive on predicting $Y$ while being maximally compressive on $X$. Therefore, following the standard information bottleneck theory \cite{tishby2000information}, our objective function is denoted as follows: 
\begin{eqnarray}
\label{eq:information_bottlenect}
\max_{Z} I(Z;Y)-\beta \cdot I(Z;X),
\end{eqnarray}

where the definitions of $X$ and $Z=X^{\text{mask}}$ are given in Equation \ref{eq:x_mask}. $Y$ is the target output, $I(\cdot;\cdot)$ denotes the mutual information, and $\beta \in \mathbb{R_+}$ is a coefficient that balances the two terms in the information bottleneck loss function. The formula of mutual information $I$ should include the parameters $\theta$ which need to be optimized. For simplicity, we ignore $\theta$ in the following formulae. 

However, computing the mutual information in Equation \ref{eq:information_bottlenect} is usually computationally challenging. Therefore, we adopted variational inference method to construct a lower bound for Equation \ref{eq:information_bottlenect}.
After constructing the lower bound and applying the reparameterization trick \cite{kingma2013auto}, we can optimize the objective utilizing stochastic gradient descent. In this subsection, we simply listed the lower bound of Equation \ref{eq:information_bottlenect}. The complete details on the derivation of lower bound 
for variational contextual sentence masking is shown in Appendix \ref{appendix:vcsm}.

Assuming that the true joint distribution is $P(X, Y, Z)$ and $X, Y, Z$ are random variables which have the following conditional dependency: $Y \leftrightarrow X \leftrightarrow Z$. And $x, y, z$ are instances of random variables.
The lower bound of Equation \ref{eq:information_bottlenect} is as follows: 
\begin{flalign}
\label{eq:izx_iyz_lower_bound}
\nonumber
\nonumber
&\sum_{x,y,z}P_{X}(x)P_{Y|X}(y|x) P_{Z|X}(z|x) \log Q_{Y|X}(y|z) 
\\
&-\beta \sum_{z,x} P_{X}(x) P_{Z|X}(z|x)\log\frac{P_{Z|X}(z|x)}{Q_{Z}(z)}
\end{flalign}
To compute Equation \ref{eq:izx_iyz_lower_bound}, we use the empirical data distribution including two Delta functions to approximate the $P_{X,Y}(x, y)$. Therefore we have the loss function of variational information bottleneck (VAB) as follows:
\begin{flalign}
\label{eq:loss_information_bottleneck}
\nonumber
\ell_{vib} = -(&\mathbb{E}_{P_{X,Y}(x, y)}[ \mathbb{E}_{P_{Z|X}(z|x)}[\log(Q_{Y|Z}(y|z)]\\&-\beta \cdot \text{KL}[P_{Z|X}(z|x)||Q_{Z}(z)]])
\end{flalign}

\subsection{Consistency Training based on Variational Contextual Sentence Masking}

In this work, we utilized a particular consistency training setting where the masked input $x^{\text{mask}}$ is generated by applying variational contextual sentence masking mentioned above on each input $x$, which can be written as follows: $x^{\text{mask}} = M \cdot x$, to improve both the interpretability and prediction performance. 
 
More specifically, inspired by \citet{xie2019unsupervised}, we propose to substitute the traditional noise injection methods with our Contextual Sentence Masking module to generate the masked input $x^{\text{mask}}$ given each input $x$ in the unlabeled dataset which can be written as follows: $x^{\text{mask}} = M \cdot x$. 
We also use the information bottleneck framework in the consistency training. The only change is to replace the ground truth label $y_i$ with the prediction $\hat{y}_i$ given the original research paper $x_i$ as the input. To be noted, the sentence mask layer is not used when predicting $\hat{y}_i$.

\subsection{Variational Information Bottleneck (VAB) Loss Function}

As shown in Figure \ref{fig:cssm_architecture}, our VAB loss functions contains two parts: a supervised VAB loss $\ell_{su}$ and an unsupervised VAB loss $\ell_{un}$. The same model is optimized in both losses.

\paragraph{Supervised VAB Loss} Since we have ground truth labels in the labeled dataset, the supervised VAB loss $\ell_{su}$ is the same as the VAB loss $\ell_{vlb}$ in Equation \ref{eq:loss_information_bottleneck} and it is denoted as follows:
\begin{flalign}
\label{eq:loss_su_ce}
\nonumber
\ell_{su} = -(&\mathbb{E}_{P_{X,Y}(x, y)}[ \mathbb{E}_{P_{Z|X}(z|x)}[\log(Q_{Y|Z}(y|z)]\\&- \beta \cdot \text{KL}[P_{Z|X}(z|x)||Q_{Z}(z)]])
\end{flalign}
where $P_{X,Y}(x, y)$ refers to empirical distribution of complete observations.

\paragraph{Unsupervised VAB Loss}
 
As for the unsupervised VAB loss, the only difference from the supervised one is to replace the ground truth label $y$ by the prediction $\hat{y}=f(x)$ given the original research paper $x$ as the input  and and it is denoted as follows:
\begin{flalign}
\label{eq:loss_consistency_training}
\nonumber
{\ell}_{un} =  -(&\mathbb{E}_{P_{X}(x)}[\mathbb{E}_{P_{Z|X}(z|x)}[\log(Q_{Y|Z}(\hat{y}|z)]\\&-\beta \cdot \text{KL}[P_{Z|X}(z|x)||Q_{Z}(z)]])
\end{flalign}

where $P_{X}(x)$ refers to empirical distribution of incomplete observations.

\paragraph{Total Loss}
In summary, our full training objective $\ell$ can be written as follows: 
\begin{equation}
\label{eq:loss_consistency_training}
\ell = \ell_{su} + \alpha \cdot \ell_{un}
\end{equation}
where $\alpha > 0$ is a balancing hyper parameter about these two items of losses. Our goal is to minimize the full training objective $\ell$.

\section{Experimental Setup}
The proposed VCCSM method is evaluated with two typical neural network models commonly used on text classification tasks, LSTM \cite{hochreiter1997long} and BERT \cite{devlin2018bert} on two datasets. 

\subsection{Datasets}

\paragraph{RRP Dataset}
RRP dataset is proposed by \citet{luo2020research}. RRP dataset contains 399 labeled and 2,170 unlabeled research articles in social science fields.
In this paper, randomly selected 300 (150:1;150:0) labeled and 2,170 unlabeled samples are treated as the training dataset. The remaining 99 (51:1;48:0) labeled research articles are considered as the testing dataset. More details about the RRP dataset 
are shown in Appendix \ref{appendix:details_datasets}. PDFMiner \cite{shinyama2014pdfminer} is used to extract the text in the raw pdf files for both labeled and unlabeled datasets. 
Therefore, the text format of labeled and unlabeled datasets are the same. 

\paragraph{ECHR Dataset} European Convention of Human Rights (ECHR) \citep{chalkidis2019neural} is a publicly available English legal judgment prediction dataset which contains 11,478 cases. Each case has a list of paragraphs describing the facts. The task is to predict whether one given case is judged as violated or not. The ECHR dataset is split into training, development, and testing datasets with the number of cases of 7,100, 1,380 and 2,998. The average number of tokens for training, development, and testing datasets are 2,421, 1,931, and 2,588, respectively.

\subsection{Implementation Details}
The LSTM model we used has a bidirectional hidden layer, and it's initialized with 300-dimensional google's pre-trained word embeddings. We fix the embedding layer and update other parameters in LSTM to achieve the best performance. As for BERT model, a published BERT pre-trained model (``bert-base-uncased''\footnote{https://huggingface.co/bert-base-uncased}) is utilized as the embedding layer of LSTM model. 
We first use our corpus to pre-train the BERT model and then fine-tune it in the VCCSM classifier's training. In each epoch, the model is first trained on labeled data, followed by unlabeled data. The hidden state of the [CLS] token of the last layer is considered as the sentence representation.

Because the average length (words) of all the documents in the labeled and unlabeled datasets is about 10,000, we set the the maximum length of words in our paper to 10,000. Since VCCSM method is sentence masking and we need to split the text of research paper into sentences. We use period, question mark, and semicolon to conduct the splitting. After some statistical analysis, the average length (words) of each sentence is around 25.
For a fair comparison with word masking method, we set the maximum length of sentences in each document to 400. It means that we set the maximum length of words in each document to 10,000 in all models. In the experiments, for RRP dataset, the number of labeled and unlabeled datasets are 4,00 and 2,170 research papers respectively. As for ECHR dataset, 2,000 cases in the training dataset are considered as the labeled and the remaining 5,100 cases as the unlabeled.

\subsection{Interpretability Metrics}

\subsubsection{AOPC}

The first interpretability metric we used is area over the perturbation curve (AOPC) \cite{samek2016evaluating,nguyen2018comparing} which is obtained by computing the average change of prediction probability by deleting top $n$ important words and it can evaluate the model interpretablity on faithness.
Since our proposed VCCSM is sentence masking method, we calculate the average change of prediction probability by deleting top $n$ key sentences in the explanations of the papers. Therefore, AOPC used in our paper is defined as follows: 
\resizebox{\linewidth}{!}{
\begin{minipage}{\linewidth}
\vspace{-0.3cm}
\begin{flalign}
\nonumber
\text{AOPC}(f) = \frac{1}{T+1} \sum_{i=1}^{T} \left(f(x_i)-f(x_i \backslash\{s_1,...,s_n\})\right),
\end{flalign}
\end{minipage}
}
where $f(x_i \backslash\{s_1,...,s_n\})$ is the probability for the predicted class on the $i_{\text{th}}$ document in RRP when the top $n$ sentences on importance are removed. Higher AOPC score is better.

\subsubsection{Post-hoc Accuracy}

The second interpretability metric utilized in this paper is post-hoc accuracy metric \cite{chen2018learning} which is computed by counting how many testing examples' predictions are changed by utilizing only extracted top $n$ words to classify. For our VCCSM models, we used top $n$ key sentences. The formula to calculate the post-hoc accuracy in our paper is as follows:

\begin{minipage}{0.95\linewidth}
\vspace{-0.4cm}
\begin{flalign}
\nonumber
\text{Acc}_{p}(f,n) = \frac{1}{T} \sum_{i=1}^{T}\mathbbm{1}[f(\{s_1,...,s_n\})=f(x_i)],
\end{flalign}
\end{minipage}

where $T$ is the number of examples in the testing dataset, $\{s_1,...,s_n\}$ are the top $n$ sentences on importance in the $i_{\text{th}}$ document.
Higher post-hoc accuracy is better.

\section{Results}
We tested our proposed models on two text classification datasets (RRP along with ECHR), and the details about prediction accuracy and interpretability are described in this section.

\begin{table*}[!ht]
\footnotesize
\centering
\begin{tabular}{c|c|c|c|c|c|c}
\hline
\multicolumn{1}{l|}{}&
\multicolumn{3}{c|}{RRP}&
\multicolumn{3}{c}{ECHR}
\\
Methods                                 & Acc      &AOPC & Post-hoc & Acc      &AOPC & Post-hoc\\ \hline\hline
LSTM Word Masking \cite{chen2020learning}      & 60.61\%  & 11.16\%   & 50.51\% & 84.86\% & 10.32\% & 65.84\% \\ 
BERT's Attention Weights (words) & 64.65\% & 11.70\%     & 60.61\% & 84.26\% & 15.06\% & 73.75\%\\
BERT Word Masking \cite{chen2020learning}      & 65.66\% & 12.05\%     & 61.62\%  & 85.06\% & 16.30\% & 76.38\% \\ 
SOTA Extractive Summarization \cite{cui2021sliding}  & 65.66\% & 12.86\%     & 57.58\% & 85.39\% & 19.57\% & 75.52\% \\
BERT's Attention Weights (sentences)  & 65.66\% & 13.62\%     & 62.63\% & 85.39\% & 22.61\% & 81.49\% \\\hline\hline
LSTM Sentence Masking + Contextual + Consistency       & 65.66\% & 22.19\%     & 63.64\% & 86.06\% & 30.53\% & 84.22\% \\
BERT Sentence Masking + Contextual + Consistency      &  \textbf{68.69\%} & \textbf{24.02\%}     & \textbf{65.66\%} & \textbf{87.66\%} & \textbf{32.78\%} & \textbf{86.59\%} \\ \hline\hline
\end{tabular}
\caption{Comparison between VCCSM and other methods on testing accuracy, area over the perturbation curve (AOPC), and post-hoc accuracy on RRP and ECHR datasets.
}
\label{tab:experimental_result_accuracy_othermethods}
\end{table*}

\begin{table*}[!ht]
\small
\centering
\begin{tabular}{ccccc}
\hline
Model  & Methods                                 & Accuracy      &AOPC & Post-hoc \\ \hline\hline
 ~ & Proposed LSTM VCCSM       & 65.66\% & 22.19\%     & 63.64\% \\
 \multirow{1}{*}{LSTM}                 &  w/o consistency training       & 62.63\%  & 14.29\%   & 60.61\% \\
  ~                    & w/o contextual masking       & 63.64\%  & 19.10\%   & 62.63\% \\
 \hline
 ~                     & Proposed BERT VCCSM       & 68.69\% & 24.02\%     & 65.66\% \\ 
 \multirow{1}{*}{BERT} & w/o consistency training        & 65.66\% & 16.38\%     & 62.63\% \\ 
 ~                     & w/o contextual masking       &  66.67\% & 21.16\%     & 64.65\% \\ \hline
\end{tabular}
\caption{Ablation study of proposed VCCSM (LSTM \& BERT Sentence Masking + Contextual + Consistency) on testing accuracy, area over the perturbation curve (AOPC), and post-hoc accuracy on RRP dataset.}
\label{tab:ablation_study}
\end{table*}

\subsection{Quantitative Evaluation}
We evaluate the interpretability of VCCSM model against other types of models via the AOPC \cite{samek2016evaluating,nguyen2018comparing} and post-hoc accuracy \cite{chen2018learning} metrics. We also listed the performance with varying number of the unlabeled data in Appendix \ref{appendix:unlabeled_scale} and it shows that the performance become higher with more unlabeled data.

\begin{figure*}[!h]
\centering
{
\includegraphics[scale=0.59,angle=270 ]{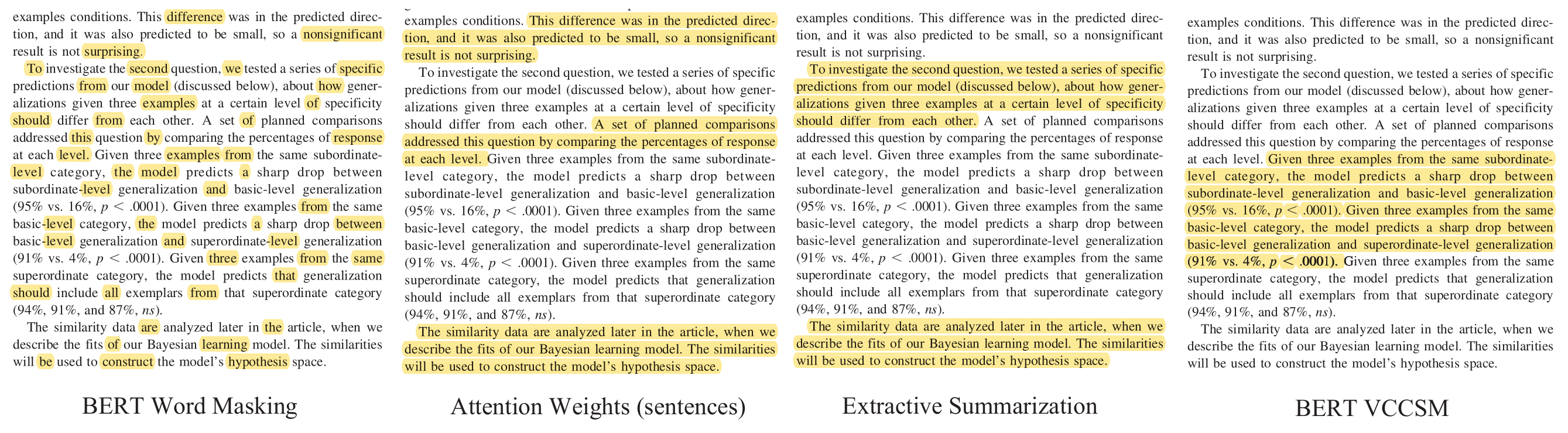}
}
\caption{Highlighted explanations (words or sentences) of BERT word masking, attention weights (sentences), SOTA extractive summarization, and BERT VCCSM methods for a paragraph in one replicable research paper ``Word Learning as Bayesian Inference'' in Psychological Review.}
\label{fig:example_qualitative_evaluation_lstm}
\vspace{-0.44cm}
\end{figure*}

Table \ref{tab:experimental_result_accuracy_othermethods} shows the results of VCCSM (LSTM \& BERT) and other interpretable models on the RPP and ECHR datasets with top 500 words (word based methods) or 20 sentences (sentence based methods). Simialr results are obtained with varying number of sentences. For BERT's attention weights model, we extracted the words' attention weights of all heads in the last layer and average them. As for BERT's attention weights (sentences), we average the words' averaged weights in each sentence as its sentence representation. 
Extractive summarization models can also extract the key sentences for each document. In this section, we used the recent extractive summarization method \cite{cui2021sliding} as the baseline. We conduct the training on arXiv + PubMed \cite{cohan2018discourse} and our labeled + unlabeled datasets (the abstract are the summary). Training on arXiv + PubMed aims to generalize the model and make the model extract a more comprehensive of information instead of only abstract in the research paper. We can observe that our proposed models perform better than other methods in both interpretability and prediction performance on both RRP and ECHR datasets.

\paragraph{Ablation Study} In order to validate different modules in our proposed VCCSM method, we conduct the ablation study on the RRP dataset as shown in Table \ref{tab:ablation_study}. We observe the drop after removing contextual masking or consistency training (on the unlabeled data) which shows that each component benefit to the model. It is noting that we observe a larger drop on both accuracy and two interpreatability metrics without the consistency training on the unlabeled data which demonstrates that consistency training contributes more to the model.

\subsection{Qualitative Evaluation}

In this section, we conduct the qualitative evaluations and compare the explanations of different models intuitively by highlighting the words or sentences. Specifically, we draw on the Open Science pratices (e.g., mentioning how to access the data) as indicators of high reproducibility, because these practices are proposed as solutions to the reproducibility crisis in the science community \cite{simonsohn2015better,foster2017open,brodeur2020methods,dienlin2021agenda,markowitz2021tracing}. Some of those indicators which are easier to check are listed as below: (1) Publish materials, data, and code; (2) Preregister studies and submit the reports; (3) Conduct the replications by themselves; (4) Collaborate with others; (5) P-value\footnote{Probability of obtaining test results at least as extreme as the results actually observed, under the assumption that the null hypothesis is correct.} is close to 0.5.

We conduct the case studies on the testing dataset and find that our proposed methods can highlight more sentences which are related to the indicators mentioned above. A case study is shown in Figure \ref{fig:example_qualitative_evaluation_lstm}. More specifically, Figure \ref{fig:example_qualitative_evaluation_lstm} shows highlighted explanations (words or sentences) of BERT word masking, attention weights (sentences), SOTA extractive summarization, and BERT VCCSM methods for a paragraph in one replicable research paper ``Word Learning as Bayesian Inference'' \cite{xu2007word} in Psychological Review. In this case study, we extracted top 200 sentences or 5,000 words (only for BERT word masking method) but only show one paragraph highlighted results. Although all the methods provide the correct prediction, our VCCSM highlights the sentences which are related to the indicators described above. It is noting that the highlight words of BERT word masking is not so readable for the long research paper. Attention weights (sentences) and SOTA extractive summarization methods can provide informational sentences but the highlighted sentence are not related to the indicators described above. BERT VCSSM can highlight $p$-value sentences which are related to the indicators mentioned above.

\subsection{Discussion on Plausibility of Predicting Research Replicability using Text}

By looking into RRP's labeled dataset and conducting the cases studies carefully such as in Figure \ref{fig:example_qualitative_evaluation_lstm}, we discuss on whether classifying results in a research paper as replicable using text is actually sufficient to replicate the results, which is the central premise this paper is based on. Non-replicability of scientific studies largely results from unscientific, unethical research practices (e.g., p-hacking, selective reporting, data manipulation). Such practices can be manifested in the texts of research papers such as the reports of p-values, experimental procedures, etc. Generally speaking, the more problematic practices a research paper involves, the less likely its findings are valid, and the less likely it will be reproduced. Hence, by modeling the replicability of research paper with regard to its textual components that are potentially linked with the problematic practices, we can classify whether a research paper can be replicated and identify the focal sentences relevant to the prediction.

\section{Concluding Remarks}
In this paper, we proposed VCCSM
to improve both interpretability and prediction accuracy on RRP along with ECHR datasets, 
using largely unlabeled datasets. We tested VCCSM with two different neural text classifiers (LSTM and BERT) and evaluated both 
prediction accuracy and interpretability metrics.
As future work, we plan to explore other advanced interpretable models and weakly supervised methods to further improve the prediction performance and interpretability of long document classification tasks.

\section*{Acknowledgments}

This research is partially supported by National Science Foundation (NSF) under Grant No. IIS-2007951 and the Defense Advanced Research Projects Agency (DARPA) and Space and Naval Warfare Systems Center Pacific (SSC Pacific) under Contract No. N66001-19-C-4014. 
The authors would like to thank Dr. Bingjie Liu for her valuable comments on the manuscript. We also thank all the anonymous reviewers for their insightful comments.

\section{Broader Impacts}

Our paper proposed VCCSM method to build an interpretable model for long document datasets such as RRP and ECHR. Our model can provide the explanations about why a research paper is predicted as replicable or non-replicalbe and why a case is judged as violated or not so that the prediction results obtained by neural text classifier are more reliable and trustworthy. However, sometimes, our proposed methods can be misused. For example, people may try to adversarially write the new text in a research paper to fool the research replication prediction tool when they can obtain the explanations by using our interpretable models. Therefore, the proposed methods in this paper should be used with careful consideration of its potential misusing when deployed in the real-world.

\bibliography{acl_latex}

\begin{thebibliography}{50}
\expandafter\ifx\csname natexlab\endcsname\relax\def\natexlab#1{#1}\fi

\bibitem[{Alemi et~al.(2016)Alemi, Fischer, Dillon, and Murphy}]{alemi2016deep}
Alexander~A Alemi, Ian Fischer, Joshua~V Dillon, and Kevin Murphy. 2016.
\newblock Deep variational information bottleneck.
\newblock \emph{arXiv preprint arXiv:1612.00410}.

\bibitem[{Altmejd et~al.(2019)Altmejd, Dreber, Forsell, Huber, Imai,
  Johannesson, Kirchler, Nave, and Camerer}]{altmejd2019predicting}
Adam Altmejd, Anna Dreber, Eskil Forsell, Juergen Huber, Taisuke Imai, Magnus
  Johannesson, Michael Kirchler, Gideon Nave, and Colin Camerer. 2019.
\newblock Predicting the replicability of social science lab experiments.
\newblock \emph{PloS one}, 14(12).

\bibitem[{Berthelot et~al.(2019)Berthelot, Carlini, Goodfellow, Papernot,
  Oliver, and Raffel}]{berthelot2019mixmatch}
David Berthelot, Nicholas Carlini, Ian Goodfellow, Nicolas Papernot, Avital
  Oliver, and Colin Raffel. 2019.
\newblock Mixmatch: A holistic approach to semi-supervised learning.
\newblock \emph{arXiv preprint arXiv:1905.02249}.

\bibitem[{Brodeur et~al.(2020)Brodeur, Cook, and Heyes}]{brodeur2020methods}
Abel Brodeur, Nikolai Cook, and Anthony Heyes. 2020.
\newblock Methods matter: P-hacking and publication bias in causal analysis in
  economics.
\newblock \emph{American Economic Review}, 110(11):3634--60.

\bibitem[{Camerer et~al.(2016)Camerer, Dreber, Forsell, Ho, Huber, Johannesson,
  Kirchler, Almenberg, Altmejd, Chan et~al.}]{camerer2016evaluating}
Colin~F Camerer, Anna Dreber, Eskil Forsell, Teck-Hua Ho, J{\"u}rgen Huber,
  Magnus Johannesson, Michael Kirchler, Johan Almenberg, Adam Altmejd, Taizan
  Chan, et~al. 2016.
\newblock Evaluating replicability of laboratory experiments in economics.
\newblock \emph{Science}, 351(6280):1433--1436.

\bibitem[{Camerer et~al.(2018)Camerer, Dreber, Holzmeister, Ho, Huber,
  Johannesson, Kirchler, Nave, Nosek, Pfeiffer et~al.}]{camerer2018evaluating}
Colin~F Camerer, Anna Dreber, Felix Holzmeister, Teck-Hua Ho, J{\"u}rgen Huber,
  Magnus Johannesson, Michael Kirchler, Gideon Nave, Brian~A Nosek, Thomas
  Pfeiffer, et~al. 2018.
\newblock Evaluating the replicability of social science experiments in nature
  and science between 2010 and 2015.
\newblock \emph{Nature Human Behaviour}, 2(9):637--644.

\bibitem[{Chalkidis et~al.(2019)Chalkidis, Androutsopoulos, and
  Aletras}]{chalkidis2019neural}
Ilias Chalkidis, Ion Androutsopoulos, and Nikolaos Aletras. 2019.
\newblock Neural legal judgment prediction in english.
\newblock \emph{arXiv preprint arXiv:1906.02059}.

\bibitem[{Chen et~al.(2021)Chen, Feng, Ganhotra, Wan, Gunasekara, Joshi, and
  Ji}]{chen2021explaining}
Hanjie Chen, Song Feng, Jatin Ganhotra, Hui Wan, Chulaka Gunasekara, Sachindra
  Joshi, and Yangfeng Ji. 2021.
\newblock Explaining neural network predictions on sentence pairs via learning
  word-group masks.
\newblock \emph{arXiv preprint arXiv:2104.04488}.

\bibitem[{Chen and Ji(2020)}]{chen2020learning}
Hanjie Chen and Yangfeng Ji. 2020.
\newblock Learning variational word masks to improve the interpretability of
  neural text classifiers.
\newblock \emph{arXiv preprint arXiv:2010.00667}.

\bibitem[{Chen et~al.(2020)Chen, Yang, and Yang}]{chen2020mixtext}
Jiaao Chen, Zichao Yang, and Diyi Yang. 2020.
\newblock Mixtext: Linguistically-informed interpolation of hidden space for
  semi-supervised text classification.
\newblock \emph{arXiv preprint arXiv:2004.12239}.

\bibitem[{Chen et~al.(2018)Chen, Song, Wainwright, and
  Jordan}]{chen2018learning}
Jianbo Chen, Le~Song, Martin Wainwright, and Michael Jordan. 2018.
\newblock Learning to explain: An information-theoretic perspective on model
  interpretation.
\newblock In \emph{International Conference on Machine Learning}, pages
  883--892. PMLR.

\bibitem[{Clark et~al.(2018)Clark, Luong, Manning, and Le}]{clark2018semi}
Kevin Clark, Minh-Thang Luong, Christopher~D Manning, and Quoc~V Le. 2018.
\newblock Semi-supervised sequence modeling with cross-view training.
\newblock \emph{arXiv preprint arXiv:1809.08370}.

\bibitem[{Cohan et~al.(2018)Cohan, Dernoncourt, Kim, Bui, Kim, Chang, and
  Goharian}]{cohan2018discourse}
Arman Cohan, Franck Dernoncourt, Doo~Soon Kim, Trung Bui, Seokhwan Kim, Walter
  Chang, and Nazli Goharian. 2018.
\newblock A discourse-aware attention model for abstractive summarization of
  long documents.
\newblock \emph{arXiv preprint arXiv:1804.05685}.

\bibitem[{Collaboration(2012)}]{open2012open}
Open~Science Collaboration. 2012.
\newblock An open, large-scale, collaborative effort to estimate the
  reproducibility of psychological science.
\newblock \emph{Perspectives on Psychological Science}, 7(6):657--660.

\bibitem[{Collaboration et~al.(2015)}]{open2015estimating}
Open~Science Collaboration et~al. 2015.
\newblock Estimating the reproducibility of psychological science.
\newblock \emph{Science}, 349(6251):aac4716.

\bibitem[{Cui and Hu(2021)}]{cui2021sliding}
Peng Cui and Le~Hu. 2021.
\newblock Sliding selector network with dynamic memory for extractive
  summarization of long documents.
\newblock In \emph{Proceedings of the 2021 Conference of the North American
  Chapter of the Association for Computational Linguistics: Human Language
  Technologies}, pages 5881--5891.

\bibitem[{Devlin et~al.(2018)Devlin, Chang, Lee, and
  Toutanova}]{devlin2018bert}
Jacob Devlin, Ming-Wei Chang, Kenton Lee, and Kristina Toutanova. 2018.
\newblock Bert: Pre-training of deep bidirectional transformers for language
  understanding.
\newblock \emph{arXiv preprint arXiv:1810.04805}.

\bibitem[{Dienlin et~al.(2021)Dienlin, Johannes, Bowman, Masur, Engesser,
  K{\"u}mpel, Lukito, Bier, Zhang, Johnson et~al.}]{dienlin2021agenda}
Tobias Dienlin, Niklas Johannes, Nicholas~David Bowman, Philipp~K Masur, Sven
  Engesser, Anna~Sophie K{\"u}mpel, Josephine Lukito, Lindsey~M Bier, Renwen
  Zhang, Benjamin~K Johnson, et~al. 2021.
\newblock An agenda for open science in communication.
\newblock \emph{Journal of Communication}, 71(1):1--26.

\bibitem[{Dreber et~al.(2019)Dreber, Pfeiffer, Forsell, Viganola, Johannesson,
  Chen, Wilson, Nosek, and Almenberg}]{dreber2018predicting}
A~Dreber, T~Pfeiffer, E~Forsell, D~Viganola, M~Johannesson, Y~Chen, B~Wilson,
  BA~Nosek, and J~Almenberg. 2019.
\newblock Predicting replication outcomes in the many labs 2 study.
\newblock \emph{Journal of Economic Psychology}.

\bibitem[{Ebersole et~al.(2016)Ebersole, Atherton, Belanger, Skulborstad,
  Allen, Banks, Baranski, Bernstein, Bonfiglio, Boucher
  et~al.}]{ebersole2016many}
Charles~R Ebersole, Olivia~E Atherton, Aimee~L Belanger, Hayley~M Skulborstad,
  Jill~M Allen, Jonathan~B Banks, Erica Baranski, Michael~J Bernstein, Diane~BV
  Bonfiglio, Leanne Boucher, et~al. 2016.
\newblock Many labs 3: Evaluating participant pool quality across the academic
  semester via replication.
\newblock \emph{Journal of Experimental Social Psychology}, 67:68--82.

\bibitem[{Foster and Deardorff(2017)}]{foster2017open}
Erin~D Foster and Ariel Deardorff. 2017.
\newblock Open science framework (osf).
\newblock \emph{Journal of the Medical Library Association: JMLA}, 105(2):203.

\bibitem[{Han et~al.(2020)Han, Wallace, and Tsvetkov}]{han2020explaining}
Xiaochuang Han, Byron~C Wallace, and Yulia Tsvetkov. 2020.
\newblock Explaining black box predictions and unveiling data artifacts through
  influence functions.
\newblock \emph{arXiv preprint arXiv:2005.06676}.

\bibitem[{Hechtlinger(2016)}]{hechtlinger2016interpretation}
Yotam Hechtlinger. 2016.
\newblock Interpretation of prediction models using the input gradient.
\newblock \emph{arXiv preprint arXiv:1611.07634}.

\bibitem[{Hochreiter and Schmidhuber(1997)}]{hochreiter1997long}
Sepp Hochreiter and J{\"u}rgen Schmidhuber. 1997.
\newblock Long short-term memory.
\newblock \emph{Neural computation}, 9(8):1735--1780.

\bibitem[{Kingma and Welling(2013)}]{kingma2013auto}
Diederik~P Kingma and Max Welling. 2013.
\newblock Auto-encoding variational bayes.
\newblock \emph{arXiv preprint arXiv:1312.6114}.

\bibitem[{Klein et~al.(2014)Klein, Ratliff, Vianello, Adams~Jr, Bahn{\'\i}k,
  Bernstein, Bocian, Brandt, Brooks, Brumbaugh et~al.}]{klein2014investigating}
Richard~A Klein, Kate~A Ratliff, Michelangelo Vianello, Reginald~B Adams~Jr,
  {\v{S}}t{\v{e}}p{\'a}n Bahn{\'\i}k, Michael~J Bernstein, Konrad Bocian,
  Mark~J Brandt, Beach Brooks, Claudia~Chloe Brumbaugh, et~al. 2014.
\newblock Investigating variation in replicability.
\newblock \emph{Social psychology}.

\bibitem[{Klein et~al.(2018)Klein, Vianello, Hasselman, Adams, Adams~Jr, Alper,
  Aveyard, Axt, Babalola, Bahn{\'\i}k et~al.}]{klein2018many}
Richard~A Klein, Michelangelo Vianello, Fred Hasselman, Byron~G Adams,
  Reginald~B Adams~Jr, Sinan Alper, Mark Aveyard, Jordan~R Axt, Mayowa~T
  Babalola, {\v{S}}t{\v{e}}p{\'a}n Bahn{\'\i}k, et~al. 2018.
\newblock Many labs 2: Investigating variation in replicability across samples
  and settings.
\newblock \emph{Advances in Methods and Practices in Psychological Science},
  1(4):443--490.

\bibitem[{Koh and Liang(2017)}]{koh2017understanding}
Pang~Wei Koh and Percy Liang. 2017.
\newblock Understanding black-box predictions via influence functions.
\newblock In \emph{International Conference on Machine Learning}, pages
  1885--1894. PMLR.

\bibitem[{Laine and Aila(2016)}]{laine2016temporal}
Samuli Laine and Timo Aila. 2016.
\newblock Temporal ensembling for semi-supervised learning.
\newblock \emph{arXiv preprint arXiv:1610.02242}.

\bibitem[{Luo et~al.(2020)Luo, Li, Wang, and Liu}]{luo2020research}
Tianyi Luo, Xingyu Li, Hainan Wang, and Yang Liu. 2020.
\newblock Research replication prediction using weakly supervised learning.
\newblock In \emph{Proceedings of the 2020 Conference on Empirical Methods in
  Natural Language Processing: Findings}, pages 1464--1474.

\bibitem[{Markowitz et~al.(2021)Markowitz, Song, and
  Taylor}]{markowitz2021tracing}
David~M Markowitz, Hyunjin Song, and Samuel~Hardman Taylor. 2021.
\newblock Tracing the adoption and effects of open science in communication
  research.
\newblock \emph{Journal of Communication}.

\bibitem[{Miyato et~al.(2018)Miyato, Maeda, Koyama, and
  Ishii}]{miyato2018virtual}
Takeru Miyato, Shin-ichi Maeda, Masanori Koyama, and Shin Ishii. 2018.
\newblock Virtual adversarial training: a regularization method for supervised
  and semi-supervised learning.
\newblock \emph{IEEE transactions on pattern analysis and machine
  intelligence}, 41(8):1979--1993.

\bibitem[{Murdoch et~al.(2018)Murdoch, Liu, and Yu}]{murdoch2018beyond}
W~James Murdoch, Peter~J Liu, and Bin Yu. 2018.
\newblock Beyond word importance: Contextual decomposition to extract
  interactions from lstms.
\newblock \emph{arXiv preprint arXiv:1801.05453}.

\bibitem[{Nguyen(2018)}]{nguyen2018comparing}
Dong Nguyen. 2018.
\newblock Comparing automatic and human evaluation of local explanations for
  text classification.
\newblock In \emph{Proceedings of the 2018 Conference of the North American
  Chapter of the Association for Computational Linguistics: Human Language
  Technologies, Volume 1 (Long Papers)}, pages 1069--1078.

\bibitem[{Pashler et~al.(2019)Pashler, Spellman, Kang, and
  Holcombe}]{pashler2019psychfiledrawer}
H~Pashler, B~Spellman, S~Kang, and A~Holcombe. 2019.
\newblock Psychfiledrawer: archive of replication attempts in experimental
  psychology.
\newblock \emph{Online< http://psychfiledrawer. org/view\_ article\_list. php}.

\bibitem[{Sajjadi et~al.(2016)Sajjadi, Javanmardi, and
  Tasdizen}]{sajjadi2016regularization}
Mehdi Sajjadi, Mehran Javanmardi, and Tolga Tasdizen. 2016.
\newblock Regularization with stochastic transformations and perturbations for
  deep semi-supervised learning.
\newblock \emph{Advances in neural information processing systems},
  29:1163--1171.

\bibitem[{Samek et~al.(2016)Samek, Binder, Montavon, Lapuschkin, and
  M{\"u}ller}]{samek2016evaluating}
Wojciech Samek, Alexander Binder, Gr{\'e}goire Montavon, Sebastian Lapuschkin,
  and Klaus-Robert M{\"u}ller. 2016.
\newblock Evaluating the visualization of what a deep neural network has
  learned.
\newblock \emph{IEEE transactions on neural networks and learning systems},
  28(11):2660--2673.

\bibitem[{Serrano and Smith(2019)}]{serrano2019attention}
Sofia Serrano and Noah~A Smith. 2019.
\newblock Is attention interpretable?
\newblock \emph{arXiv preprint arXiv:1906.03731}.

\bibitem[{Shinyama(2014)}]{shinyama2014pdfminer}
Yusuke Shinyama. 2014.
\newblock Pdfminer.

\bibitem[{Simons et~al.(2014)Simons, Holcombe, and
  Spellman}]{simons2014introduction}
Daniel~J Simons, Alex~O Holcombe, and Barbara~A Spellman. 2014.
\newblock An introduction to registered replication reports at perspectives on
  psychological science.
\newblock \emph{Perspectives on Psychological Science}, 9(5):552--555.

\bibitem[{Simonsohn et~al.(2015)Simonsohn, Simmons, and
  Nelson}]{simonsohn2015better}
Uri Simonsohn, Joseph~P Simmons, and Leif~D Nelson. 2015.
\newblock Better p-curves: Making p-curve analysis more robust to errors,
  fraud, and ambitious p-hacking, a reply to ulrich and miller (2015).

\bibitem[{Simonyan et~al.(2014)Simonyan, Vedaldi, and
  Zisserman}]{simonyan2014deep}
Karen Simonyan, Andrea Vedaldi, and Andrew Zisserman. 2014.
\newblock Deep inside convolutional networks: Visualising image classification
  models and saliency maps.
\newblock In \emph{In Workshop at International Conference on Learning
  Representations}. Citeseer.

\bibitem[{Singh et~al.(2018)Singh, Murdoch, and Yu}]{singh2018hierarchical}
Chandan Singh, W~James Murdoch, and Bin Yu. 2018.
\newblock Hierarchical interpretations for neural network predictions.
\newblock \emph{arXiv preprint arXiv:1806.05337}.

\bibitem[{Smilkov et~al.(2017)Smilkov, Thorat, Kim, Vi{\'e}gas, and
  Wattenberg}]{smilkov2017smoothgrad}
Daniel Smilkov, Nikhil Thorat, Been Kim, Fernanda Vi{\'e}gas, and Martin
  Wattenberg. 2017.
\newblock Smoothgrad: removing noise by adding noise.
\newblock \emph{arXiv preprint arXiv:1706.03825}.

\bibitem[{Sundararajan et~al.(2017)Sundararajan, Taly, and
  Yan}]{sundararajan2017axiomatic}
Mukund Sundararajan, Ankur Taly, and Qiqi Yan. 2017.
\newblock Axiomatic attribution for deep networks.
\newblock In \emph{International Conference on Machine Learning}, pages
  3319--3328. PMLR.

\bibitem[{Tarvainen and Valpola(2017)}]{tarvainen2017mean}
Antti Tarvainen and Harri Valpola. 2017.
\newblock Mean teachers are better role models: Weight-averaged consistency
  targets improve semi-supervised deep learning results.
\newblock \emph{arXiv preprint arXiv:1703.01780}.

\bibitem[{Tishby et~al.(2000)Tishby, Pereira, and
  Bialek}]{tishby2000information}
Naftali Tishby, Fernando~C Pereira, and William Bialek. 2000.
\newblock The information bottleneck method.
\newblock \emph{arXiv preprint physics/0004057}.

\bibitem[{Xie et~al.(2019)Xie, Dai, Hovy, Luong, and Le}]{xie2019unsupervised}
Qizhe Xie, Zihang Dai, Eduard Hovy, Minh-Thang Luong, and Quoc~V Le. 2019.
\newblock Unsupervised data augmentation for consistency training.
\newblock \emph{arXiv preprint arXiv:1904.12848}.

\bibitem[{Xu and Tenenbaum(2007)}]{xu2007word}
Fei Xu and Joshua~B Tenenbaum. 2007.
\newblock Word learning as bayesian inference.
\newblock \emph{Psychological review}, 114(2):245.

\bibitem[{Yang(2018)}]{Yang2019}
Yang Yang. 2018.
\newblock The replicability of scientific findings using human and machine
  intelligence.
\newblock \url{https://www.metascience2019.org/presentations/yang-yang/}
  Metascience 2019.

\end{thebibliography}
\bibliographystyle{acl_natbib}

\appendix

\section{Appendix}

\subsection{Detailed Derivation of Lower Bound for Variational Contextual Sentence Masking in Section 4.2}\label{appendix:vcsm}

In this section, we provided the complete details on the derivation of lower bound for variational contextual sentence masking in Section 4.2.

Assuming that the true joint distribution is $P(X, Y, Z)$ and $X, Y, Z$ are random variables which have the following conditional dependency: $Y \leftrightarrow X \leftrightarrow Z$. And $x, y, z$ are instances of ramdom variables. We can have
\begin{flalign}
\label{eq:depend_x_y_z}
\nonumber
P(X,Y,Z)&=P(Z|X,Y)P(Y|X)P(X)\\
&=P(Z|X)P(Y|X)P(X).
\end{flalign}
According to the definition of $I(Z;Y)$, we have 
\begin{flalign}
\label{eq:z_y_lower_bound}
\nonumber
I(Z;Y) &= \sum_{z,y}P_{Z,Y}(z,y)\log\frac{P_{Z,Y}(z,y)}{P_{Z}(z)P_{Y}(y)}\\
&=\sum_{z,y}P_{Z,Y}(z,y)\log\frac{P_{Y|Z}(y|z)}{P_{Y}(y)}.
\end{flalign}
And we also have 
\begin{flalign}
\label{eq:p_y_z}
\nonumber
P_{Y|Z}(y|z)&=\sum_{x} P_{X,Y|Z}(x,y|z)
\\
\nonumber
&=\sum_{x} P_{Y|X}(y|x)P_{X|Z}(x|z) 
\\
&=\sum_{x}\frac{P_{Y|X}(y|x)P_{Z|X}(z|x)P_{X}(x)}{P_{Z}(z)}.
\end{flalign}
Since $P(Y|Z)$ can be intractable, $Q(Y|Z)$ is considered as a variational approximation to $P(Y|Z)$. $Q(Y|Z)$ is our decoder and a neural network. Because the Kullback Leibler divergence is non-negative, we have
\begin{flalign}
\label{eq:p_kl_y_z}
\nonumber
&~~~~~~~~~~~~~~~~~~~\text{KL}[P(Y|Z)||Q(Y|Z)] \geq 0 
\\
&\Rightarrow \sum_{y} p(y|z) \log p(y|z) \geq \sum_{y} p(y|z) \log q(y|z).
\end{flalign} Therefore, we can obtain the lower bound of $I(Z;Y)$ as follows: 
\begin{flalign}
\label{eq:iyz_lower_bound}
\nonumber
I(Z;Y&) \geq \sum_{z,y} P_{Z,Y}(z,y)\log\frac{Q_{Y,Z}(y|z)}{P_{Y}(y)}
\\&=\sum_{z,y} P_{Z,Y}(z,y)\log Q_{Y|Z}(y|z)+H(Y).
\end{flalign}
where $H(Y)=-\sum_{y}P_{Y}(y)\log P_{Y}(y)$ is entropy. According to Equation \ref{eq:depend_x_y_z}, we have
\begin{flalign}
\label{eq:rewrite_pxy_lower_bound}
\nonumber
P(Y|Z) &= \sum_{x}P_{X,Y,Z}(x,y,z)
\\
\nonumber
&=\sum_{x}P_{X,Y,Z}(x,y,z)\\&=\sum_{x} P_{X}(x) P_{Y|X}(y|x) P_{Z|X}(z|x).
\end{flalign}
Hence, we obtain the lower bound of $I(Z,Y)$ as follows:
\begin{flalign}
\nonumber
\sum_{x,y,z}P_{X}(x)P_{Y|X}(y|x)P_{Z|X}(z|x)\log Q_{Y|Z}(y|z).
\end{flalign}
As for $I(Z;X)$, similar to Equation \ref{eq:z_y_lower_bound} in the derivation of $I(Z;Y)$, we first obtain 
\begin{flalign}
\label{eq:z_y_lower_bound_2}
\nonumber
I(Z;X) &= \sum_{z,x}P_{Z,X}(z,x)\log\frac{P_{Z|X}(z|x)}{P_{Z}(z)}\\
\nonumber
&=\sum_{z,x}P_{Z,X}(z,x)\log P_{Z|X}(z|x) \\&~~~~~- \sum_{z}P_{Z}(z)\log P_{Z}(z).
\end{flalign}

\begin{figure*}[!ht]
\centering
\subfigure[LSTM Sentence Masking + Contextual + Consistency]
{
    \begin{minipage}[]{0.48\textwidth}
    \centering
    \includegraphics[width=1.04\columnwidth,height=6.1cm]{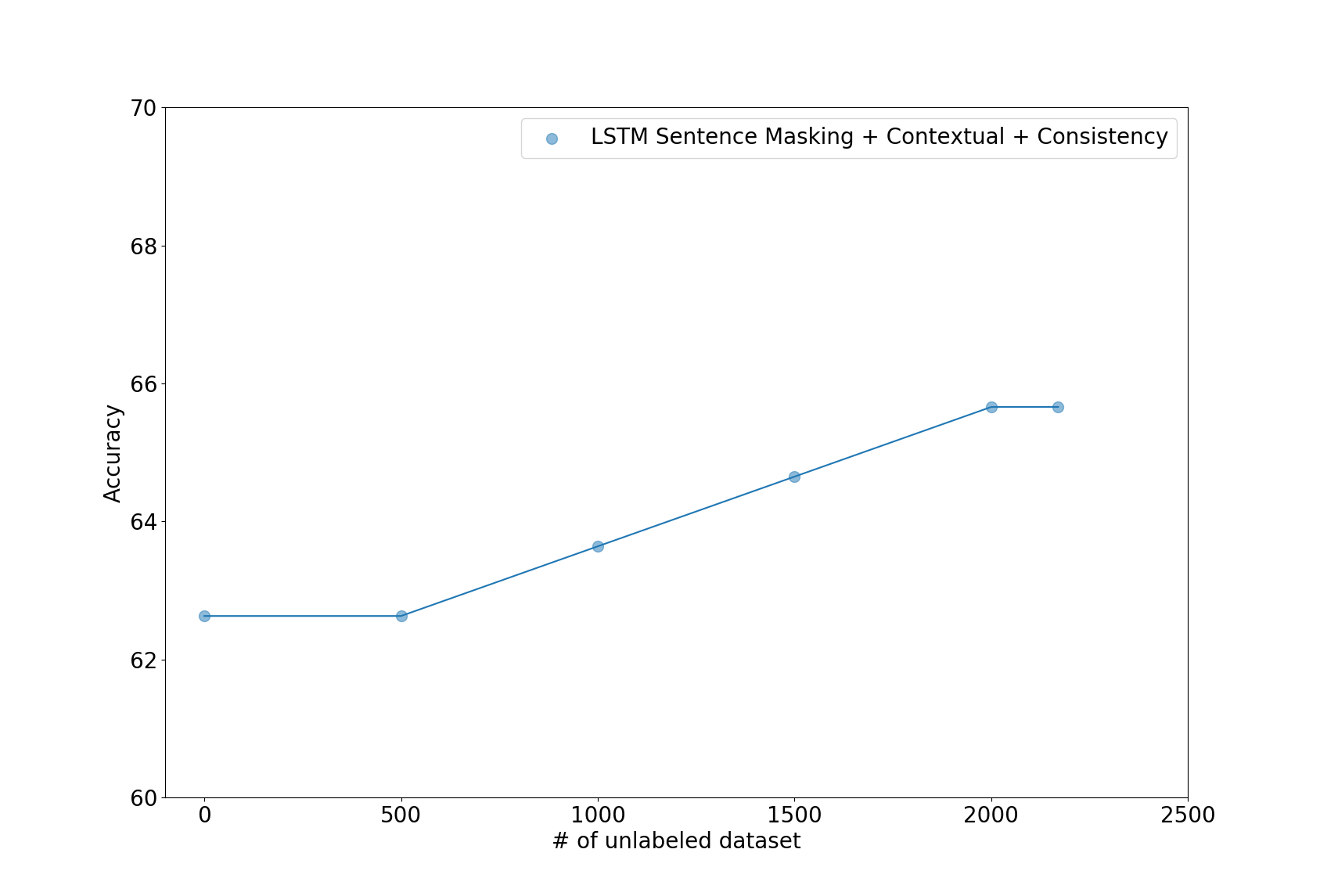} 
    \end{minipage}
}
\subfigure[BERT Sentence Masking + Contextual + Consistency] 
{
    \begin{minipage}[]{0.48\textwidth}
    \centering
    \includegraphics[width=1.04\columnwidth,height=6.1cm]{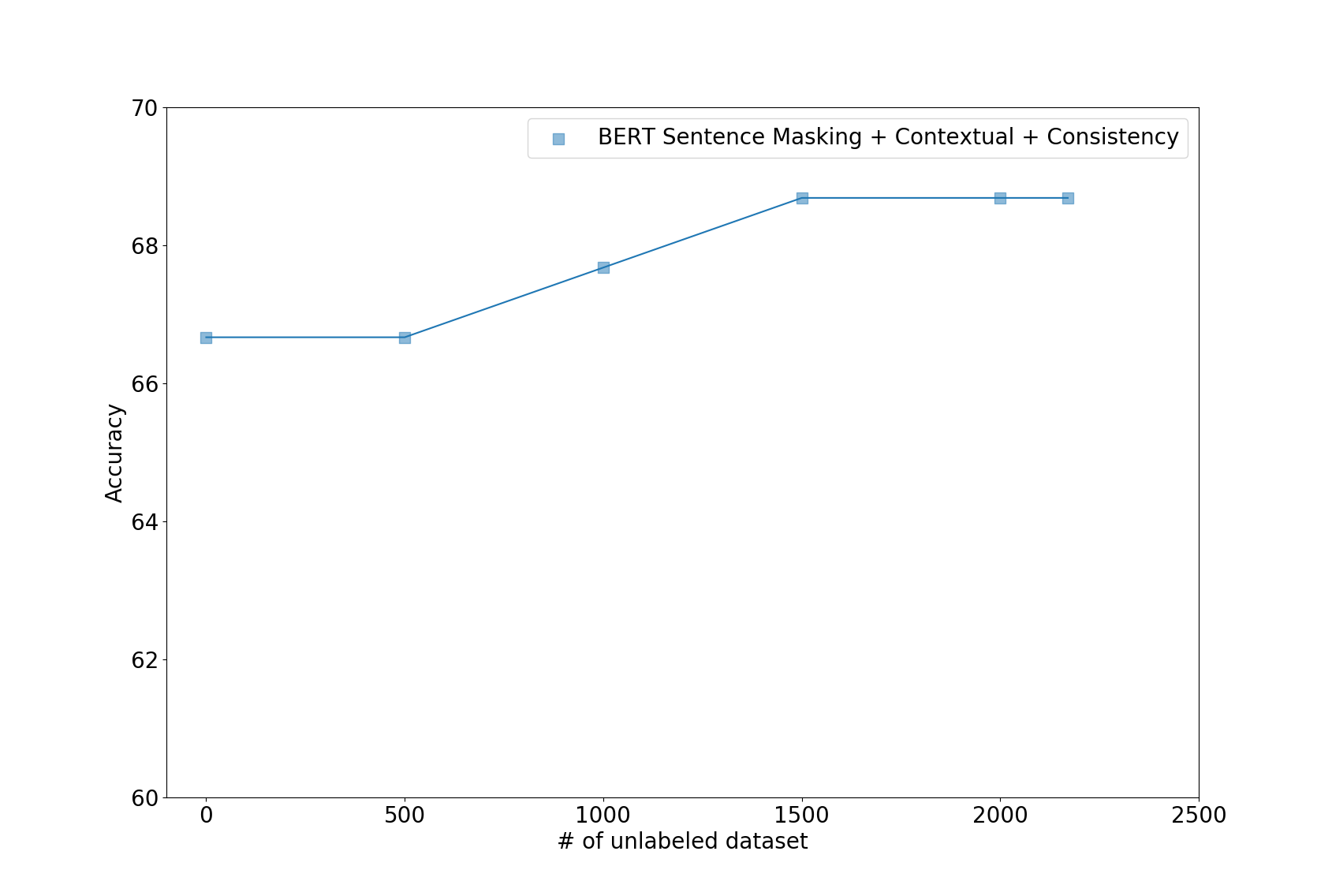}
    \end{minipage}
}
\caption{\small{Testing accuracy (\%) on RRP dataset with varying number of unlabeled dataset for VCCSM applied on two neural text classifiers (LSTM and BERT)}}
\label{fig:fig_la_ha}
\end{figure*}

Because the marginal distribtuion of $Z$, $P(Z)=\sum_{x} P_{Z|X}(z|x) P_{X}(x)$ in which the computation might be difficult, we replace $P(Z)$ by a variational approximation of $Q(Z)$. Since $\text{KL}[P(Z)||Q(Z)] \geq 0 \Rightarrow \sum_{z} P_{Z}(z)\log P_{Z}(z) \geq \sum_{z} P_{Z}(z) \log Q_{Z}(z)$, we can get the upper bound of $I(Z;X)$ as follows:
\begin{flalign}
\label{eq:izx_upper_bound}
\nonumber
I(Z;X) &\leq \sum_{z,x}P_{Z,X}(z,x)\log P_{Z|X}(z|x) \\
\nonumber
&~~~~~- \sum_{z,x}P_{Z,X}(z,x)\log Q_{Z}(z)
\\ &\leq \sum_{z,x} P_{X}(x)P_{Z|X}(z|x)\log\frac{P_{Z|X}(z|x)}{Q_{Z}(z)}.
\end{flalign}
Combining Equation \ref{eq:iyz_lower_bound} and \ref{eq:izx_upper_bound}, we can get the lower bound of $I(Z;Y)-\beta I(Z;X)$ as follows:
\begin{flalign}
\label{eq:izx_iyz_lower_bound}
\nonumber
&\sum_{x,y,z}P_{X}(x)P_{Y|X}(y|x)P_{Z|X}(z|x) \log Q_{Y|X}(y|z) 
\\
\nonumber
&-\beta \sum_{z,x} P_{X}(x)P_{Z|X}(z|x)\log\frac{P_{Z|X}(z|x)}{Q_{Z}(z)}.
\end{flalign}

\subsection{Details of RRP Dataset}\label{appendix:details_datasets}

In the RRP dataset proposed by \citet{luo2020research}, the labeled datset are collected from eight research replication projects which are the  Registered Replication Report (RRR) \citep{simons2014introduction}, Many Labs 1 \citep{klein2014investigating}, Many Labs 2 \citep{klein2018many}, Many Labs 3 \citep{ebersole2016many}, Social Sciences Replication Project (SSRP)  \citep{camerer2018evaluating}, PsychFileDrawer \citep{pashler2019psychfiledrawer}, Experimental Economics Replication Project \cite{camerer2016evaluating}, and Reproducibility Project: Psychology (RPP) \citep{open2012open}. Among 399 labeled data in the RRP dataset, 201 are labeled as `1' (replicable) and the remain 198 are annotated as `0' (non-replicable). We observe that the labeled data in the RRP dataset is balanced.

In addition, RRP dataset also contains 2,170 research articles as the unlabeled dataset. \citet{luo2020research} observed that most papers in the labeled dataset in the RRP dataset are economical and psychology related. Among those papers, they are mainly from American Economic Review and Psychological Science journals. Therefore, a python crawler is written by \citet{luo2020research} to get 2,170 published research articles on the American Economic Review (Jan 2011 - Dec 2014) and Psychological Science websites (Jan 2006 - Dec 2012). The number of articles crawled from American Economic Review and Psychological Science websites are 981 and 1,189 respectively.

\subsection{Performance with Varying Number of Unlabeled Data}\label{appendix:unlabeled_scale}

We conducted the experiments to test our model's effectiveness by varying number of unlabeled data for VCCSM applied on two neural text classifiers (LSTM and BERT). From Figure \ref{fig:fig_la_ha}, we can observe that, with more unlabeled data, the testing accuracy become higher on both LSTM Sentence Masking + Contextual + Consistency and BERT Sentence Masking + Contextual + Consistency models, which validates the effectiveness of using unlabeled data.

\end{document}